\pdfoutput=1

\documentclass[11pt]{article}

\usepackage[final]{acl}

\usepackage{times}
\usepackage{latexsym}

\usepackage[T1]{fontenc}

\usepackage[utf8]{inputenc}

\usepackage{microtype}

\usepackage{inconsolata}

\usepackage{svg}
\usepackage{graphicx}
\usepackage{xcolor}
\usepackage{colortbl}
\usepackage{color}
\usepackage{bbding}
\usepackage{adjustbox}
\usepackage{booktabs}
\usepackage{makecell}
\usepackage{booktabs}
\usepackage{threeparttable}
\usepackage{multirow}
\usepackage{enumitem}
\usepackage{float}
\usepackage{subfig}
\usepackage{overpic}
\usepackage{amsmath}
\usepackage{cleveref}
\usepackage{arydshln}
\usepackage{makecell}
\definecolor{BLUE}{HTML}{7091A9}
\definecolor{GREEN}{HTML}{66938A}
\definecolor{BROWN}{HTML}{A59D6F}
\definecolor{GR}{HTML}{0CA01C}
\definecolor{RD}{HTML}{ff0000}
\definecolor{ALG}{HTML}{008080}

\usepackage[normalem]{ulem}
\usepackage[ruled,linesnumbered,commentsnumbered]{algorithm2e}
\useunder{\uline}{\ul}{}
\title{CDT: A Comprehensive Capability Framework for Large Language Models Across \underline{C}ognition, \underline{D}omain, and \underline{T}ask}

\author{
 Haosi Mo$^{1}$,
 Xinyu Ma$^{1}$,
 Xuebo Liu$^{1}$\thanks{~~Corresponding Author},
 Derek F. Wong$^{2}$,
 Yu Li$^{3}$,
 Jie Liu$^{4}$, and
 Min Zhang$^{1}$
 \\
    \textsuperscript{\rm1}Institute of Computing and Intelligence, Harbin Institute of Technology, Shenzhen, China \\
    \textsuperscript{\rm2}NLP$^{2}$CT Lab, Department of Computer and Information Science, University of Macau \\
    \textsuperscript{\rm3}School of Integrated Circuits, Zhejiang University, Hangzhou, China\\
    \textsuperscript{\rm4}State Key Lab of Smart Farm Technologies and Systems, Harbin Institute of Technology\\
    \texttt{\{alessamo0411,mxinyuma,yu.li.sallylee\}@gmail.com}, \texttt{derekfw@um.edu.mo} \\
    \texttt{\{liuxuebo,jieliu,zhangmin2021\}@hit.edu.cn}
    }

\begin{document}
\maketitle
\begin{abstract}
Recent advances in Large Language Models (LLMs) have significantly enhanced their capabilities, highlighting the need for comprehensive evaluation frameworks that extend beyond task-specific benchmarks. 
However, existing benchmarks often focus on isolated abilities, lacking a holistic framework for assessing LLM capabilities. 
To address this gap, we propose the \textbf{C}ognition-\textbf{D}omain-\textbf{T}ask (CDT) framework, which comprehensively measures a model's capabilities across three dimensions. 
We expand the scope of model capability definitions at the cognitive level by incorporating the Cattell-Horn-Carroll cognitive theory, refining the categorization of model capabilities. 
We apply CDT in two directions: dataset capability evaluation and data selection. Experiments show that our capability metrics correlate well with downstream performance and can support effective dataset analysis and construction. The experiments on data selection also show significant improvements in both general and specific benchmarks, achieving scores of 44.3 and 45.4, with an increase of 1.6 and 2.2 points over the baselines, respectively. These results validate the effectiveness and practicality of CDT. Source code and models are available at \url{https://github.com/Alessa-mo/CDT}.
\end{abstract}

\section{Introduction}
Recent advances in Large Language Models (LLMs) have significantly expanded their capabilities. The introduction of reinforcement learning~\citep{kumar2024traininglanguagemodelsselfcorrect, wang-etal-2024-math, hu2024enablingintelligentinteractionsagent} and chain-of-thought reasoning~\citep{wei2023chainofthoughtpromptingelicitsreasoning, wang2023selfconsistencyimproveschainthought} has further enhanced their reasoning abilities. Notable LLMs such as OpenAI's o1~\citep{o1} and DeepSeek R1~\citep{deepseekai2025deepseekr1incentivizingreasoningcapability} have demonstrated remarkable reasoning capabilities. As LLMs become more sophisticated, accurately evaluating their underlying abilities is increasingly crucial. Current benchmarks, such as MMLU~\citep{hendrycks2021measuringmassivemultitasklanguage}, AlpacaEval~\citep{dubois2024lengthcontrolledalpacaevalsimpleway}, and GSM8K~\citep{cobbe2021trainingverifierssolvemath}, are widely used to assess these capabilities. 

However, many of them focus on isolated aspects of model capabilities, such as coding, commonsense reasoning, or specific task performance, and the ability dimensions are always task-oriented and limited, without a holistic framework that systematically categorizes and defines the full spectrum of LLM capabilities. For instance, benchmarks like MMLU evaluate knowledge mastery across academic disciplines but overlook dimensions like code generation. Recent efforts like FLASK~\citep{ye2024flaskfinegrainedlanguagemodel} and FAC$^2$E~\citep{wang-etal-2024-fac2e} focus on multi-model comparisons but fall short in capability decomposition and multi-dimensional analysis. Additionally, while works like INSTAG~\citep{lu2024instag} explore capability applications, definitions remain underdeveloped. Those works raise the fundamental question: \textit{What core capabilities are essential for an effective large language model?} 

\begin{table}[!tp]
\centering
\setlength{\tabcolsep}{3pt} 
\renewcommand{\arraystretch}{0.95} 
\scalebox{0.6}{
\begin{tabular}{lcccccc}
\toprule
\textbf{\thead{Framework}} &
  \textbf{\thead{Open Source \\ Tagging Models}} & 
  \textbf{\thead{Multiple \\ Dimensions}} & 
  \textbf{\thead{Capability \\ Decomposition}} & 
  \textbf{\thead{Cognition \\ Oriented}} & 
  \textbf{\thead{Domain \\ Oriented}} & 
  \textbf{\thead{Task \\ Oriented}} \\ \midrule
\footnotesize \textbf{FLASK}         & \textcolor{red}{\XSolidBrush} & \textcolor{GR}{\Checkmark} & \textcolor{red}{\XSolidBrush} & \textcolor{GR}{\Checkmark} & \textcolor{GR}{\Checkmark} & \textcolor{red}{\XSolidBrush}  \\ 
\footnotesize \textbf{FAC$^2$E  }    & \textcolor{red}{\XSolidBrush} & \textcolor{GR}{\Checkmark} & \textcolor{GR}{\Checkmark} & \textcolor{GR}{\Checkmark} & \textcolor{red}{\XSolidBrush} & \textcolor{GR}{\Checkmark} \\ 
\footnotesize \textbf{INSTAG}        & \textcolor{GR}{\Checkmark} & \textcolor{red}{\XSolidBrush} & \textcolor{red}{\XSolidBrush} & \textcolor{red}{\XSolidBrush} & \textcolor{GR}{\Checkmark} & \textcolor{red}{\XSolidBrush} \\
\rowcolor[HTML]{c0c0c0} \footnotesize \textbf{CDT (Ours)} & \textcolor{GR}{\Checkmark} & \textcolor{GR}{\Checkmark} & \textcolor{GR}{\Checkmark} & \textcolor{GR}{\Checkmark} & \textcolor{GR}{\Checkmark} & \textcolor{GR}{\Checkmark} \\ \bottomrule
\end{tabular}
}
\caption{Comparison between our CDT framework with existing capability frameworks. ``Open Source Tagging Models'' denotes if it provides trained models for capability annotation. ``Multiple Dimensions'' reflects whether the framework supports more than one capability dimension. ``Capability Decomposition'' refers to the ability to break down complex capabilities into finer-grained sub-skills. The last three columns assess whether the framework explicitly covers cognition, domain, and task dimensions. As shown, our CDT framework addresses the gaps and limitations of existing methods across multiple dimensions.}
\label{tab:feature-compare}
\end{table}

To address this, we propose the \textbf{C}ognition-\textbf{D}omain-\textbf{T}ask (CDT) framework, a comprehensive multi-dimensional taxonomy for defining, annotating, and utilizing LLM capabilities across three dimensions: cognition, domain, and task. Our core motivation is that a comprehensive capability analysis must answer three fundamental questions for any given instruction: how to do it, which corresponds to Cognition; what it is about, which corresponds to Domain; and what to do, which corresponds to Task. By deconstructing instructions along these three orthogonal dimensions, CDT provides a holistic and systematic approach to categorizing the full spectrum of LLM capabilities. At the cognitive level, CDT incorporates Cattell-Horn-Carroll (CHC) theory~\citep{CHC}, selecting and refining 18 core cognitive abilities suited to LLM behavior. At the domain level, we identify nine domain scenarios commonly encountered by LLMs and further refine these into 33 distinct subdomains. At the task level, drawing inspiration from prior work on dataset construction~\citep{wang2022supernaturalinstructionsgeneralizationdeclarativeinstructions,wang-etal-2023-self-instruct,ouyang2022traininglanguagemodelsfollow,wang-etal-2024-inscl}, we systematically categorize task types across diverse instructions, culminating in a taxonomy of 16 task types. We conduct a comparative analysis between existing capability frameworks and our proposed CDT framework, with the results summarized in Table~\ref{tab:feature-compare}. 

After constructing the CDT framework, we extend its application to LLMs in two directions. We first conduct dataset evaluation using two metrics: \texttt{Coverage} and \texttt{Balance}, which correlate well with downstream performance, demonstrating that CDT can provide practical guidelines for capability-aware data curation in future dataset construction. Then, we apply CDT in data selection to enhance model performance, proposing a diversity-driven selection method to ensure general capability and a capability-oriented strategy which identifies the specific capabilities required by the target test sets. Across both general and specific scenarios, experiments show that our data selection methods achieve average scores of 44.3 and 45.4, with an increase of 1.6 and 2.2 points over the baselines, respectively. These results significantly outperform other capability-related methods and baseline approaches. Our \textbf{main contributions} are as follows:

\begin{itemize}[leftmargin=1em]
\item We propose CDT, a comprehensive framework that systematically categorizes LLMs' abilities across cognition, domain, and task. 
\item We develop specialized tag models for each dimension to enable fine-grained tagging of capacities at the instruction level. 
\item We explore the application of the CDT framework in dataset evaluation and data selection, which effectively reflect dataset quality and lead to notable improvements in model performance.
\item We will release all the data, tag models, and training scripts used in our CDT framework.
\end{itemize}

\section{Related Works}
\paragraph{Definitions of LLMs' Capability}
Research on defining LLM capabilities can be grouped into two approaches. The first integrates capabilities with data, optimizing learning through data distribution adjustments~\citep{nottingham2024skill,polo2025sloth,chen2024skill,wu2024mixture,commonIT,lu2024instag}. For example, \citet{chen2024skill} propose a method based on a skill set graph, where mastering one skill aids the acquisition of others, though this method is dataset-specific. Similarly, \citet{wu2024mixture} use an MLP-based scoring network for data allocation in fine-tuning, treating datasets as distinct capabilities. The second approach defines model capabilities from task- and domain-specific perspectives, often relying on labeled data for evaluation. \citet{zhong2025law} present a hierarchical framework with foundational and complex abilities, while \citet{ye2024flaskfinegrainedlanguagemodel} analyze open-source LLMs to identify four capabilities, subdividing them into 12 skills for comprehensive evaluation. While these approaches offer valuable insights, they often define capabilities in narrow ways, either focusing on isolated aspects, overlooking the underlying cognitive processes, or lacking a holistic, multi-dimensional structure. Our work addresses this gap by introducing the CDT framework, which integrates cognitive principles to systematically organize LLM capabilities across cognition, domain, and task.

\paragraph{Applications of LLMs' Capability}
Capability frameworks are often applied to develop evaluation benchmarks for large models. \citet{xia-etal-2024-fofo} introduce FoFo, which evaluates LLMs' abilities across domains based on format-following. For general evaluation, \citet{hendrycks2021measuringmassivemultitasklanguage}, \citet{dubois2024lengthcontrolledalpacaevalsimpleway}, and \citet{srivastava2022beyond} have developed broad competency benchmarks. \citet{zhong2025law} assess capabilities using prompts, while \citet{ye2024flaskfinegrainedlanguagemodel} evaluate models based on responses and instruction alignment. For domain-specific improvements, several studies have proposed different approaches: \citet{wang2024retaskrevisitingllmtasks} integrate capability frameworks with Chain-of-Thought (CoT) to enhance task-specific abilities; \citet{lee2024thanos} introduce THANOS for multi-turn dialogue; \citet{xu2023latent} present LaRS to improve reasoning by selecting data with similar capabilities; \citet{rao-etal-2025-apt} focus on enhancing weak capabilities through error-based learning; and \citet{ke2025aquilt} aim to build specialist capabilities by synthesizing high-relevance data from unlabeled text. While prior studies have extended capability frameworks in certain contexts, most still focus on evaluating LLM capabilities, with limited exploration of broader applications. To address this, we apply the proposed CDT framework to scenarios such as dataset analysis and data selection, thereby extending its utility beyond conventional evaluation.

\section{Method}
\subsection{Capability Framework Construction}\label{sec:framework-construction}
In our proposed CDT framework, we define model capabilities from three perspectives: cognition, domain, and task. The three dimensions are designed to be orthogonal, allowing for a context-dependent analysis of full instructions. A detailed discussion of the framework's design rationale is provided in \Cref{app:CDTexplain}. While the domain and task perspectives have been extensively explored in recent research, we build upon this foundation with adjustments to better capture their nuances. From the cognition perspective, we define capabilities through the lens of the CHC theory in cognitive science. The CHC theory, grounded in earlier explorations of human cognition~\citep{carroll2003higher, cattell1963theory, horn1965fluid, flanagan2000wechsler}, serves as a foundational model in cognitive science~\citep{mcgrew2004internal}. In the realm of computer science, numerous studies have demonstrated the critical role of cognitive capabilities in LLMs and artificial intelligence~\citep{zhao2022cognitive, lieto2018role, song2024m3giacognitioninspiredmultilingual}. Our overall capability framework is shown in Figure~\ref{fig:framwork}. 

\paragraph{Cognition} The CHC theory categorizes human cognitive abilities into three hierarchical levels. Stratum I consists of ``narrow'' abilities, which represent specialized skills developed through experience, learning, or the application of targeted methodologies~\citep{carroll1993human}. Stratum II encompasses ``broad'' abilities, which are more abstract and general in nature. Stratum III represents the highest level, with a single general cognitive ability acting as an overarching factor. In our framework, we focus exclusively on the Stratum I abilities defined by \citet{CHC}, as they provide specific abilities and detailed definitions that are more directly applicable than those found in the other two levels. The process of constructing LLM cognitive capabilities follows these steps:

\begin{figure}[!tp]
\centering
\includegraphics[width=\linewidth]{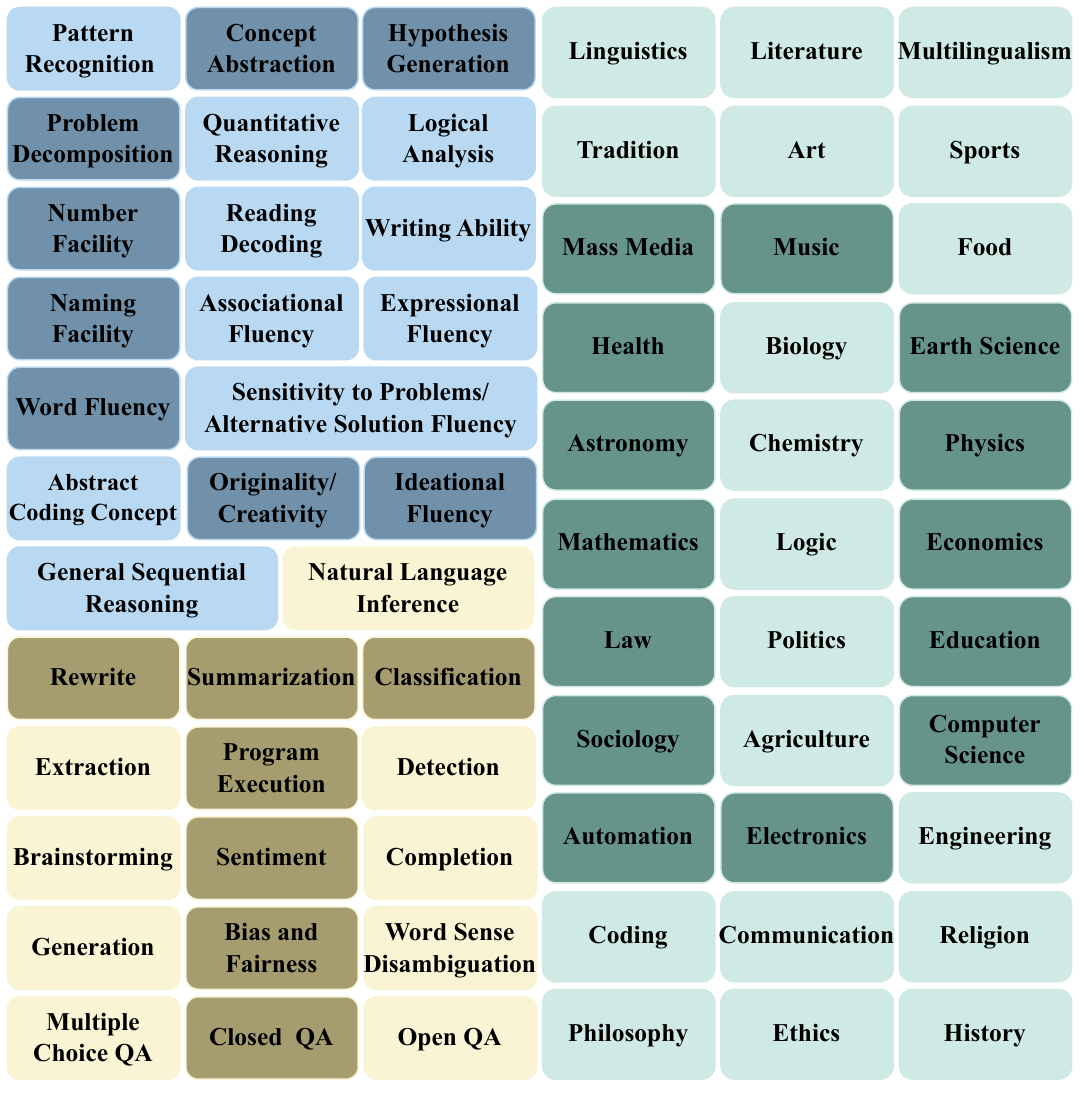}
\caption{The model capability framework we define, where the blue section represents the \textcolor{BLUE}{C}ognition dimension, the green section represents the \textcolor{GREEN}{D}omain dimension, and the brown section represents the \textcolor{BROWN}{T}ask dimension. The shaded region is used to visually emphasize our CDT framework.}
\label{fig:framwork}
\end{figure}

\begin{itemize}[leftmargin=1em]
\item \textbf{Cognition Construction}: To align with the linguistic focus of LLMs, we begin by filtering the cognitive abilities defined by CHC, which span multiple human modalities such as vision, hearing, and speech. We exclude non-linguistic abilities and domain-specific knowledge, as domain expertise is addressed separately in our framework. We also remove skills that are essential for humans but not as crucial for models, such as memory-related abilities. While this CHC-based foundation is robust, it may still overlook certain skills exhibited by LLMs. To address this, we augment the set with capabilities particularly relevant to LLMs, such as logical analysis, abstract coding concepts, and problem decomposition. After this process, the number of abilities is reduced from 82 to 16.

\item \textbf{Definition Refinement}: To better align with language models, we refine certain ability definitions. Notably, the ability Induction, originally defined as ``the ability to discover the underlying characteristic (e.g., rule, concept, process, trend, class membership) that governs a problem or a set of materials,'' often leads to ambiguity in capability tagging. Its broad and abstract nature makes it frequently assigned across diverse instructions. To address this, we subdivide it into three specific capabilities: pattern recognition, concept abstraction, and hypothesis generation. After these refinements, the total number of cognitive capabilities is $N_c = 18$, and we define the cognition dimension as $\mathcal{C} = \{c_i\}_{i=1}^{N_c}$, where each $c_i$ denotes a specific cognitive capability. The detailed cognitive capability construction procedure is provided in \Cref{app:cogselect}.
\end{itemize}
\paragraph{Domain} Based on \citet{ye2024flaskfinegrainedlanguagemodel}, which categorizes 38 domains, we construct the domain dimension of our framework. However, we observe that certain domains, such as business and marketing, exhibit considerable similarity, potentially introducing ambiguity in capability tagging models and leading to label distribution dispersion in the following process of annotating capabilities. So, we manually refine the domain set. First, we merge similar domains into one domain to reduce ambiguity. Then, we expand the domain coverage by adding underrepresented domains such as earth science and tradition. After refinement, the total number of domains is $N_d = 33$, and the domain dimension is defined as $\mathcal{D} = \{d_i\}_{i=1}^{N_d}$, where each $d_i$ represents a specific categorized domain.
\paragraph{Task} For task categorization, we draw inspiration from \citet{wang-etal-2024-inscl}, who classify the 76 tasks in SuperNI~\citep{wang2022supernaturalinstructionsgeneralizationdeclarativeinstructions} into 16 taxonomies, as well as from related work such as \citet{bach-etal-2022-promptsource} and \citet{ouyang2022traininglanguagemodelsfollow}, which offer widely accepted, fine-grained, and comprehensive categorizations. Taking task granularity and completeness into account, we ultimately categorize ${N_t}=16$ task taxonomies. For task definition, we synthesize information from Wikipedia and prior work \citep{ding-etal-2023-enhancing} to formulate detailed definitions for each task. The task dimension is defined as $\mathcal{T} = \{t_i\}_{i=1}^{N_t}$, where $t_i$ is the task we define.

Finally, the whole capability framework $\mathcal{F}$ is:
\begin{equation}
\mathcal{F} = \{ (c, d, t) \mid c \in \mathcal{C},\ d \in \mathcal{D},\ t \in \mathcal{T} \}
\end{equation}
Details on the categorization and definitions of each capability are provided in \Cref{app:tagdef}.

\subsection{Capability Tagging Model Training}
To facilitate the practical use of our framework, we train a capability tagging model for each dimension. We first prompt GPT-4o~\citep{openai2024gpt4ocard} to annotate fine-grained capability labels for each instruction in the training data due to its exceptional comprehension abilities. Given the importance of cognitive abilities in human intelligence, each data point is annotated with up to two cognitive capabilities, and one tag for both domain and task. 

We construct a training set of 49K samples from seven public instruction datasets, with 1K held out as a test set. Then we use the annotated dataset to fine-tune three annotators on the Qwen2.5-7B-Base~\citep{qwen2.5} model. To validate the performance of the trained annotators, we use the GPT-generated labels as the ground truth and evaluate the models on the test set. The accuracy rates for cognition, domain, and task tags are 93.1\%, 81.2\%, and 80.9\%, respectively, with an average score of 85.1\%, supporting the validity and reliability of the CDT tagging system. Further training details, datasets, prompt designs, human evaluation results, and a cost-benefit analysis of our annotation strategy are provided in \Cref{apx:tagging model}.

\section{CDT for Dataset Evaluation and Data Selection}
While the CDT framework offers a comprehensive definition of model capabilities, its application to LLMs remains an area requiring further exploration. Leveraging CDT's ability to classify data at the instruction level based on capabilities, we focus on two key application scenarios: evaluating the capability characteristics of existing instruction datasets and guiding the selection of training data to enhance model performance.

\subsection{Capability-Aware Dataset Evaluation}\label{sec:metrics}
To understand the quality and capability distribution of existing instruction datasets, and thereby guide future dataset construction more effectively, we introduce a capability-aware evaluation approach based on CDT. Given a labeled instruction dataset $D_i$ where each instance is annotated with composite capability triplet $(c,d,t)$, we then define the capability composites within $D_i$ as $T_i$.
\begin{equation}
    T_i = \rm Composites(\it {D_i})
\end{equation}
where $ \rm Composites$ means getting all the capability composites in a given labeled 
dataset.
We then define two quantitative metrics for dataset-level capability assessment: \texttt{Coverage} and \texttt{Balance}.
\begin{itemize}
\item \textbf{\texttt{Coverage}} measures how many distinct capability composites the dataset contains relative to the full capability space, defined as $\texttt{Coverage} = |T_i|/{|\mathcal{F}|}$.
\item \textbf{\texttt{Balance}} reflects the uniformity of the distribution over composite capabilities in the dataset. It is computed as the Shannon entropy: $\texttt{Balance} = -\sum_{t_i \in T_i} p(t_i) \log p(t_i)$, where $p(t_i)$ is the empirical probability of composite triplet $t_i$ in dataset $D_i$.
\end{itemize}
A higher \texttt{Coverage} indicates broader capability representation. This concept aligns with existing research. For example, INSTAG defines a similar metric, referred to as the unique tag coverage rate for the overall tag set, emphasizing the importance of diverse capability representation. Similarly, research by \citet{zhang2024instruction} explicitly states that the diversity of the instruction set largely determines generalization to unseen tasks, underscoring the critical role of diversity in enabling performance on novel tasks. It also points out that the uneven distribution within the training set can affect generalization ability, which in turn leads to our definition of the \texttt{Balance} metric. A higher \texttt{Balance} reflects a more uniform distribution across capabilities. This observation is echoed in other studies that stress the importance of data balance for robust model training \citep{kandpal2023large, shao-etal-2024-balanced}. Both of the metrics are desirable for building generalizable models. We employ these metrics in \Cref{exp:dataset} to evaluate a range of popular instruction datasets.

\subsection{Capability-Guided Data Selection}
Beyond supporting capability evaluation and analysis, CDT can also serve practical purposes in downstream applications. To demonstrate its effectiveness, we apply CDT to data selection scenarios for LLM instruction fine-tuning. This approach enables the systematic enhancement of training data quality and relevance, ultimately improving LLM performance on downstream tasks. 

Prior to implementing the data selection process, we first annotate the collected data pool $D_{pool}$ using the CDT framework to ensure precise capability-based categorization, resulting in the labeled dataset $D^{'}_{pool}$. As in the previous section, we define the capability composites within $D^{'}_{pool}$ as:
\begin{equation}
    T_d = \rm Composites(\it {D^{'}_{pool}})
\end{equation}
We then explore two practical strategies under this framework: a diversity-driven selection method to improve general capability coverage, and a capability-oriented filtering method to support specific scenario enhancement.
\paragraph{Diversity-Driven General Scenario Data Selection} \label{sec:general-method} When training LLMs, data diversity plays a crucial role in enhancing model performance and generalization~\citep{miranda2024beyond,NEURIPS2023_ac662d74}. Therefore, we propose a diversity-driven general data selection method based on CDT. Firstly, we define the selected training dataset as $D_{train}$ and the composite capability assigned to $D_{train}$ as $T_{s}$.
\begin{equation}
    T_s = \rm Composites(\it {D_{train}})
\end{equation}
For diversity-driven applications, our goal is to enlarge $T_{s}$ as much as possible. Then we define a threshold $R$, which denotes the ratio of $T_{s}$ to $T_{d}$. We quantify the attribute diversity as $R = |T_s|/{|T_d|}$, where $| \cdot |$ denotes the cardinality (i.e., the number of elements) of a set. 
The value of $R$ reflects the coverage rate of unique composite capabilities within the selected sub-dataset relative to the entire data pool. Our selection criterion aims to maximize the proximity of R to 1. 
Based on this, if a data point $d \in D_{pool}$ could increase $R$, we add the composite of $d$ to $T_s$ and $d$ itself to $D_{train}$ as training data. When $R$ can no longer be increased, we perform an average selection from $D_{pool}$ to fill the gaps in the capability composite of $T_s$.
\paragraph{Capability-Oriented Specific Scenario Data Selection} \label{sec:specific-method} When applying the capability framework in the capability-oriented specific scenario, we first label the validation set of the test task to obtain the labeled dataset $D_{valid}$. Then, we tag $D_{valid}$ with our annotators to form $D^{'}_{valid}$ and use the same method as in the diversity-driven approach to extract all combinations of abilities $T_v$ from $D^{'}_{valid}$.
\begin{equation}
    T_v = \rm Composites(\it {D^{'}_{valid}})
\end{equation}

We aim to perform an average selection of the data from $D^{'}_{pool}$ based on the combinations of capabilities in $T_v$. However, in practice, $T_v$ may be limited to a small subset of combinations of capabilities, and the amount of data corresponding to these combinations in $D^{'}_{pool}$ may not be sufficient to support our selection.
To address this issue, we further decompose the capabilities in $T_v$. Specifically, we break down the triplet of capabilities $f=(c,d,t)$ into binary pairs $(c,d),(c,t),(d,t)$, creating a binary set $T^{*}_{v}$, and further into individual dimensions $(c),(d),(t)$, forming a unary set $T^{\star}_{v}$.
When the triplet set $T_v$ does not yield enough data, we first perform random selection on $T^{*}_{v}$, followed by selection on $T^{\star}_{v}$ in successive stages. This approach ensures sufficient data collection while preserving the concentration of capabilities. We present the details of the two algorithms in~\Cref{apx:algorithm}.
\section{Empirical Analysis of Instruction Dataset Capabilities}\label{exp:dataset}
\paragraph{Datasets} We conduct an empirical analysis on eight publicly available instruction datasets widely used in LLM instruction tuning. These datasets include: {Chain of Thought}~\citep{wei2023chainofthoughtpromptingelicitsreasoning}, {Dolly}~\citep{DatabricksBlog2023DollyV2}, {Open Assistant}~\citep{NEURIPS2023_949f0f8f}, {Flan V2}~\citep{Longpre2023}, {WizardLM}~\citep{xu2024wizardlm}, {Alpaca-GPT4}~\citep{alpaca-gpt4}, {Self-Instruct}~\citep{wang-etal-2023-self-instruct}, and {Unnatural Instructions}~\citep{honovich2023unnatural}. Each dataset is annotated using our capability taggers to extract the composite tuples. We then compute the metrics introduced in~\Cref{sec:metrics}. Following~\citet{lu2024instag}, we collect the AlpacaEval~\citep{alpaca_eval} score for model performance.
\begin{figure}[!tp]
    \centering
    \scalebox{0.2}{\includegraphics{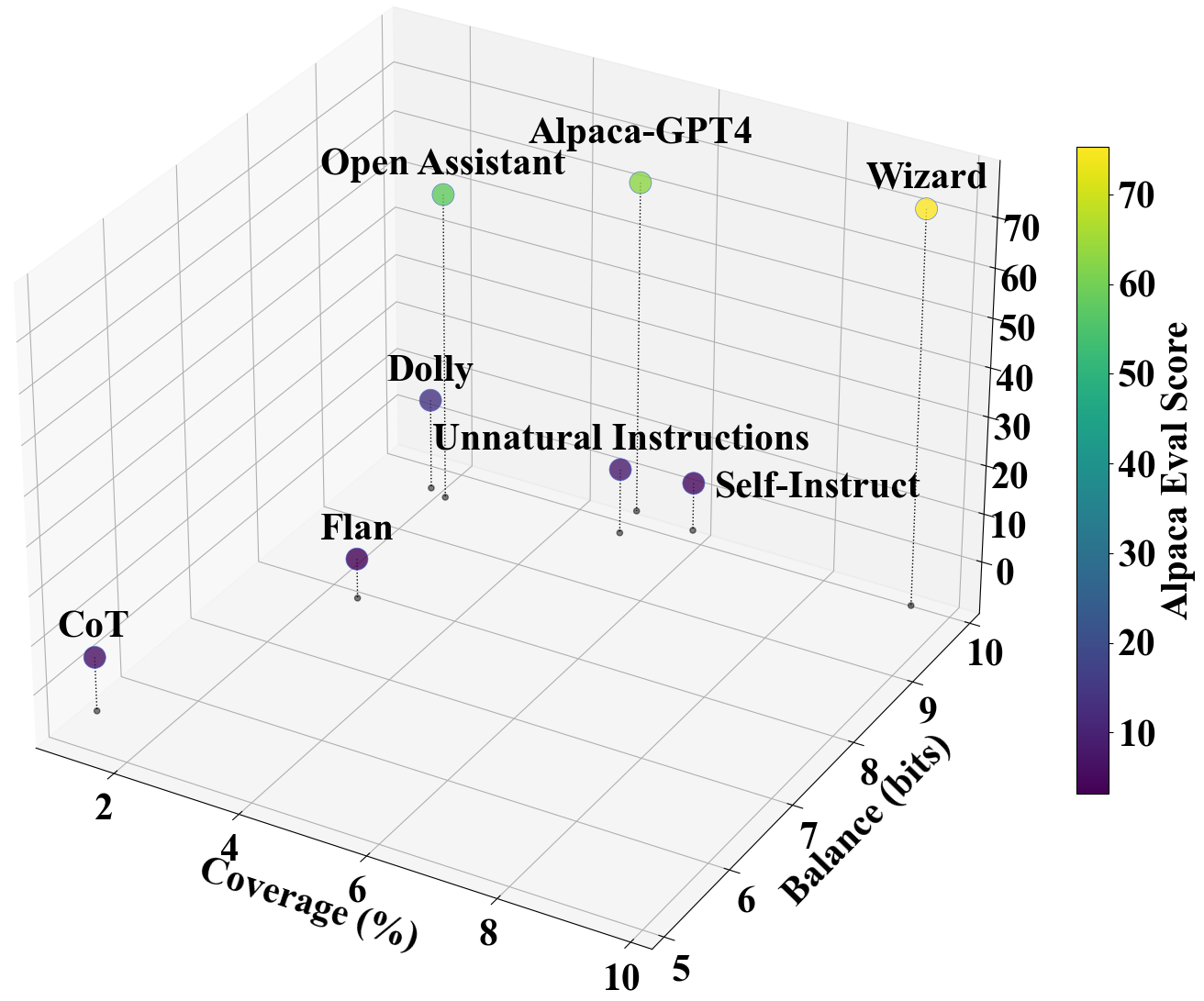}}
    \caption{Open-source instruction datasets analysis based on their capability \texttt{Coverage} and \texttt{Balance}. The Z-axis and the point color both indicate AlpacaEval score, with brighter colors corresponding to higher performance.}
    \label{fig:3d}
\end{figure}
\paragraph{Analysis} \Cref{fig:3d} visualizes the relationship between \texttt{Coverage}, \texttt{Balance}, and AlpacaEval scores for these datasets. As shown, there is a positive correlation between the two metrics and the model performance. Datasets achieving higher scores on both \texttt{Coverage} and \texttt{Balance} generally yield models with superior AlpacaEval scores. Notably, top-performing datasets such as Wizard, Alpaca-GPT4, and Open Assistant are positioned in the upper-right region of the plot, indicating high values for both our proposed metrics. Interestingly, Open Assistant, despite a moderate \texttt{Coverage} score, achieves a strong AlpacaEval score, potentially due to its exceptional \texttt{Balance} score, highlighting the crucial impact of data balance. Conversely, datasets like CoT and Flan, which score lower on these two quality indicators, correspondingly result in models with lower AlpacaEval scores. 

\paragraph{Performance Variance} We also observe that datasets with similar quantitative scores may lead to very different model performance. For example, Alpaca, Unnatural Instructions and Self-Instruct show similar metrics, but differ widely in effectiveness. We attribute this to the quality of the response annotations: Alpaca uses GPT-4-generated outputs, while the latter two rely on earlier models such as text-davinci-002/003, which are substantially weaker. Since our capability taggers operate on the instruction side, they may overlook differences in response quality, resulting in the observed discrepancy. We believe this also explains the performance gap between Open Assistant and Dolly. Although both are human-annotated, Open Assistant relies on global crowdsourcing, which results in higher-quality responses compared to Dolly, whose annotations come from Databricks employees.

Our empirical analysis demonstrates that \texttt{Coverage} and \texttt{Balance} are effective indicators of dataset quality, and that a combination of comprehensive \texttt{Coverage} and well-distributed \texttt{Balance} is crucial for training models with high performance. These findings suggest that CDT can serve as practical guidelines for capability-aware data curation in future dataset construction.

\section{Data Selection Experiments}
\label{data selection exp}
\subsection{Experiment Setup}
\paragraph{Data Pool and Base Model} To evaluate and apply our proposed capability framework, CDT, across both diversity-driven general scenario and capability-oriented specific scenario, we utilize the following datasets: (1) Aggregated high-quality datasets, including Flan V2 and CoT; and (2) Open-ended generation datasets with human-annotated responses, such as Dolly and Open Assistant. From these four datasets, we compile a pool of approximately 270K data points.
Since our annotators are trained using Qwen2.5-7B\footnote{\url{https://huggingface.co/Qwen/Qwen2.5-7B}}, we select Llama2-7B-Base\footnote{\url{https://huggingface.co/meta-llama/Llama-2-7b}} as the base model to mitigate any potential bias between the tagging model and the experimental model.
We use open-instruct\footnote{\url{https://github.com/allenai/open-instruct}} and lm-eval~\cite{eval-harness} for all tests.

\paragraph{Baselines} We conduct the following experiments for comprehensive comparison:
\begin{itemize}[leftmargin=1em]
    \item \textbf{Base}: We evaluate the pre-trained Llama2-7B-Base model on the benchmarks.
    \item \textbf{ALL}: We train the Llama2-7B-Base model using all the data from the data pool.
    \item \textbf{Random}: We randomly sample data from the data pool to train the Llama2-7B base model.
    \item \textbf{INSTAG}: \textbf{(1)} For the diversity-driven general scenario, we adopt the diversity approach outlined by~\citet{lu2024instag}, utilizing their annotation model to label the training data.\textbf{(2)} For the capability-oriented specific scenario, we use only the INSTAG annotator for tag labeling. We then average the sample data from the data pool based on the capabilities tagged in the valid set.
\end{itemize}

\paragraph{Configuration} We fine-tune the Llama2-7B-Base model using Low-Rank Adaptation (LoRA) \citep{hu2021loralowrankadaptationlarge}, specifically targeting the attention module. Distributed training is conducted using DeepSpeed~\citep{10.1145/3394486.3406703}. During training, the maximum sequence length is set to 2048, with a batch size of 64 and training epochs as 3. 

\begin{table}[!tp]
\centering
\vspace{-0.5em}
\renewcommand\tabcolsep{5.5pt}
\renewcommand\arraystretch{1.1}

\scalebox{0.62}{
\begin{tabular}{lcccccc}
\toprule
\textbf{Methods}        & \textbf{ARC-C} & \textbf{MMLU} & \textbf{BBH}  &  \textbf{CEVAL} & \textbf{TYDIQA} & \textbf{AVG.} \\ \midrule
\multicolumn{7}{c}{\textit{Baselines}}                                                                             \\ \hdashline
\textbf{Base}           &  43.5     & 45.2    & \textbf{41.6} & 31.9                 & 47.8            & 42.0          \\
\textbf{All}            & 44.5           & 45.9    & 39.6          & 35.6                 & \textbf{53.3}   & {\ul 43.8}    \\
\textbf{Random}         & 45.0  & 45.5 & {\ul 39.8}    & 32.9        & 50.4      & 42.7 \\
\textbf{InsTag}         & 44.8           & 45.8          & 39.3          & 33.2                 & 51.9            & 43.0          \\ \hdashline
\multicolumn{7}{c}{\textit{Our Methods}}                                                                           \\ \hdashline
\textbf{CDT\_Cognition} & 45.3           & 45.3          & 38.2          & {\ul 36.6}           & 51.9            & 43.5          \\
\textbf{CDT\_Domain}    & {\ul 45.9}     & {\ul 46.1}    & 38.5          & 34.3                 & 52.2            & 43.4          \\
\textbf{CDT\_Task}      & 45.7           & {\ul 46.1}    & 39.3          & 35.9                 & 50.5            & 43.5          \\
\rowcolor[HTML]{C0C0C0} 
\textbf{CDT}            & \textbf{46.1}  & \textbf{46.3} & 38.8          & \textbf{36.9}        & {\ul 53.2}      & \textbf{44.3} \\ \bottomrule
\end{tabular}}

\caption{Results of applying CDT in diversity-driven general data selection, using 20\% of the data pool for training. \textbf{Bold} indicating the best performance and \underline{underline} indicating the second-best performance.}

\label{tab:general-result}
\end{table}

\subsection{Experiments on the General Scenario} We first apply CDT to data selection in the diversity-driven general scenario. By extracting capability distributions from the data pool, we select diverse training data and evaluate performance on Llama2-7B-Base. Additional results on Mistral-7B-Base are provided in \Cref{app:mistral} to demonstrate the generalizability of our method.

\paragraph{Benchmarks} We conduct experiments using the following benchmarks:
\textbf{ARC-C}~\citep{clark2018thinksolvedquestionanswering}: We use the Challenge portion for testing, with accuracy as the evaluation metric.
\textbf{MMLU}~\citep{hendrycks2021measuringmassivemultitasklanguage}:  We report the average accuracy score under 5-shot settings.
\textbf{BBH}~\citep{srivastava2022beyond}: We use the CoT prompt and report the accuracy score.
\textbf{C-Eval}~\citep{huang2023ceval}: We use accuracy on 5-shot as the evaluation metric.
\textbf{TyDiQA}~\citep{tydiqa}: We use the GoldP task and report the average F1 score under 1-shot settings.

\paragraph{Results} As shown in Table~\ref{tab:general-result}, our method achieves the best overall performance, ranking first on most benchmarks. For BBH, all the methods exhibit performance degradation compared to the base model. We hypothesize that this may be due to a certain degree of overlap between the fine-tuning data and the model's training data, which compromises the model’s ability to generalize to complex reasoning tasks.
Furthermore, since our method considers the capabilities from three dimensions, we also conduct separate experiments for each dimension. Notably, even when using a single capability dimension, our method consistently outperforms Random and INSTAG. Among these, the approach that jointly considers all three dimensions outperforms those that consider only a single dimension across the majority of evaluation metrics. These results highlight the accuracy of our CDT framework in defining capabilities and demonstrate the effectiveness of our  diversity-driven data selection method in practice.

\begin{table}[!tb]
\centering
\vspace{-0.5em}
\renewcommand\tabcolsep{5.5pt}
\renewcommand\arraystretch{1.1}
\scalebox{0.62}{
\begin{tabular}{@{}clcccccc@{}}
\toprule
\textbf{Volume}       & \textbf{Methods} & \textbf{ARC-C} & \textbf{BBH}  & \textbf{MMLU} & \textbf{CEVAL} & \textbf{TYDIQA} & \textbf{AVG.} \\ \midrule
\multirow{2}{*}{\textbf{5\%}}  & \textbf{INSTAG}  & 44.3           & 38.3          & 44.4          & 32.1           & 49.4            & 41.7          \\
                      & \textbf{CDT}     & {\ul 45.6}     & \textbf{39.4} & 45.7          & 32.7           & 50.1            & 42.7          \\ 
                      \hdashline
\multirow{2}{*}{\textbf{20\%}} & \textbf{INSTAG}  & 44.8           & {\ul 39.3}    & 45.8          & 33.2           & {\ul 51.9}      & 43.0          \\
                      & \textbf{CDT}     & \textbf{46.1}  & 38.8          & {\ul 46.3}    & \textbf{36.9}  & \textbf{53.2}   & \textbf{44.3} \\
                      \hdashline
\multirow{2}{*}{\textbf{40\%}} & \textbf{INSTAG}  & 45.2           & \textbf{39.4} & {\ul 46.3}    & {\ul 33.7}     & 51.5            & 43.2          \\
                      & \textbf{CDT}     & 45.1           & 38.1          & \textbf{46.7} & \textbf{36.9}  & 51.6            & {\ul 43.7}           \\ \bottomrule
\end{tabular}}
\caption{The results of our method across different data selection volumes and our approach achieve the optimal results at 20\%. The results are presented with \textbf{bold} indicating the best performance and \underline{underline} indicating the second-best performance.}
\label{tab:data_vol_analysis}
\end{table}

\paragraph{Impact of Data Volume on CDT Performance}
We conduct experiments by selecting 5\%, 20\%, and 40\% of the data from the overall data pool. The results are presented in Table~\ref{tab:data_vol_analysis}. Using 20\% of the data, our method, CDT, yields the best performance compared to other volumes. However, even at these data volumes, our CDT data selection methods still outperform INSTAG in all cases. These results highlight the robustness and stability of our approach across different data volumes. Based on these findings, we choose to use the 20\% data configuration for the remaining experiments.

\begin{figure}
    \centering
    \includegraphics[width=\linewidth]{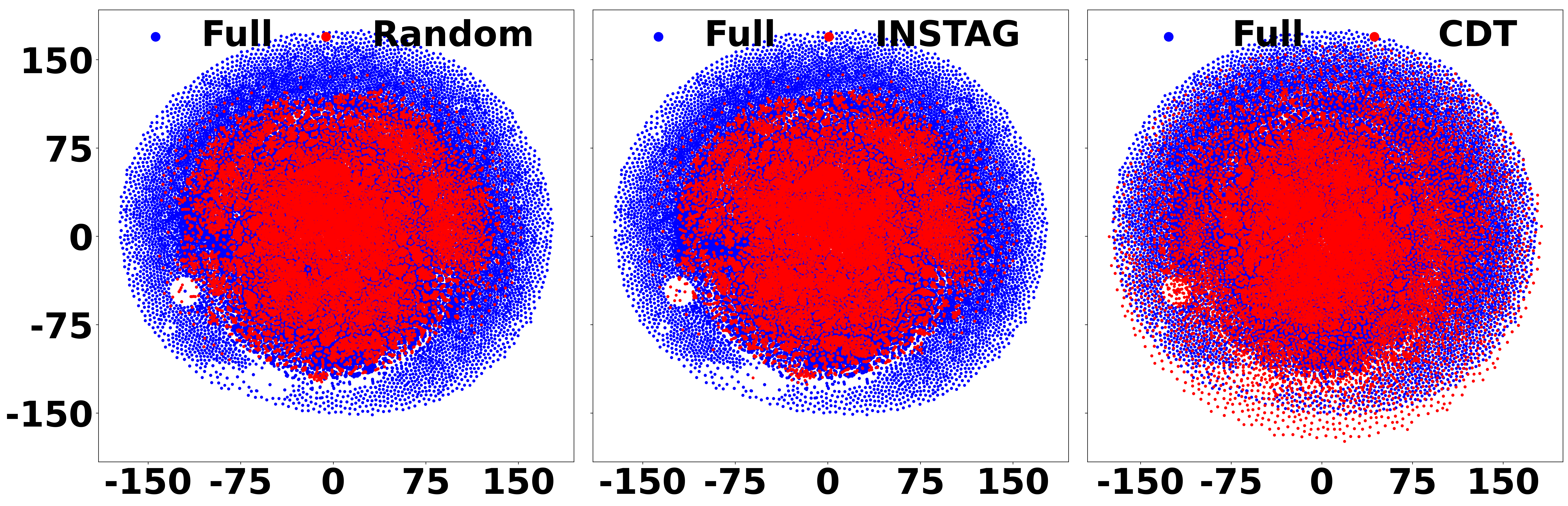}
    \caption{Diversity analysis using t-SNE on the data selected by Random, Instag, and CDT. Blue dots represent the distribution of all data, while red dots indicate the distribution of data selected by different methods.}
    \label{fig:general_diversity_analysis}
\end{figure}

\paragraph{Comparison of Data Diversity Across Methods} In the diversity-driven general scenario, training data diversity is critical to model performance. To assess the effectiveness of CDT, we analyze the diversity of data selected by Random, INSTAG, and CDT.  Following \citet{gao2024boostingmanytomanymultilingualmachine}, we use Llama2-7B-Chat to extract data representations and apply t-SNE for visualization. As shown in Figure~\ref{fig:general_diversity_analysis}, the red points of CDT are more widely dispersed than those of INSTAG and Random, indicating that the data selected by the CDT method exhibits greater diversity. This advantage in data diversity aligns with the performance improvements observed in our benchmark tests, explaining why CDT outperforms other methods in the diversity-driven general scenario. It further reinforces the rationale behind the capability definitions in our CDT framework. 

\subsection{Experiments on the Specific Scenario}
After validating CDT in the general scenario, we further test its effectiveness in the capability-oriented specific scenario, where models require data tailored to specific capabilities. Using the method introduced in~\Cref{sec:specific-method}, we conduct a detailed analysis of three target test sets.

\paragraph{Test Datasets}
We conduct experiments using the following test datasets: DROP~\citep{dua2019drop}, GSM~\citep{cobbe2021trainingverifierssolvemath}, and HISTORY, where HISTORY is a resampled subset of four history-related tasks from the MMLU benchmark. To align with the application method proposed in \Cref{sec:specific-method}, we select a maximum of 200 samples from the validation set of each task for tagging and data selection. For datasets that do not include a validation set, we randomly split 200 samples from test set to form one. If a dataset contains fewer than 200 samples, we use the full validation set available.

\begin{table}[!tp]
\centering
\vspace{-0.5em}
\renewcommand\tabcolsep{5.5pt}
\renewcommand\arraystretch{1.1}
\scalebox{0.7}{
\begin{tabular}{lccccc}
\toprule
                                   & \multicolumn{2}{c}{\textbf{\makecell{DROP}}} & \textbf{\makecell{GSM}} & \textbf{\makecell{HISTORY}} &                                 \\ \cmidrule(lr){2-3} \cmidrule(lr){4-4} \cmidrule(lr){5-5}
\multirow{-2}{*}{\textbf{Methods}} & \textbf{EM}                   & \textbf{F1}       & \textbf{EM}               & \textbf{Acc.}               & \multirow{-2}{*}{\textbf{AVG.}} \\ \midrule
\textbf{Base}                     & 0.0                      & 1.3                      & 14.5             & 51.0                           & 16.7                           \\
\textbf{All}                      & {\ul 49.0}               & \textbf{58.3}            & {\ul 21.0}         & 51.3                           & {\ul 44.9}                     \\
\textbf{Random}                   & 46.7                     & 55.8                     & 19.0               & 51.2                           & 43.2                           \\
\textbf{InsTag}                   & 47.9                     & {\ul 57.2}                     & 19.0               & {\ul 52.4}                     & 44.1                           \\
\rowcolor[HTML]{C0C0C0} 
\textbf{CDT}                       & \textbf{49.3}            & \textbf{58.3}            & \textbf{21.5}    & \textbf{52.5}                  & \textbf{45.4}    
\\ \bottomrule
\end{tabular}}
\caption{The results of using CDT for data selection in the capability-oriented specific scenario, using 20\% of the data pool for
training. The best performance is marked in \textbf{bold}, and the second-best is marked with an \underline{underline}. Our method achieves the highest performance across all three test sets.}
\label{tab:specific-result}
\end{table}

\paragraph{Result}
As shown in \Cref{tab:specific-result}, CDT consistently achieves the highest performance across all test sets. While full-data training approaches achieve similar results on DROP and GSM, our approach attains better results using only 20\% of the full dataset, demonstrating significantly improved data efficiency. Furthermore, on the HISTORY test set, the full-data baseline performs similarly to Random, yet remains 1.2 points below our approach. These results highlight the exceptional performance of CDT in capability-oriented specific scenario, demonstrating its effectiveness.

\begin{figure}
    \centering
    \subfloat[Cognition]{
        \includegraphics[width=0.3\linewidth]{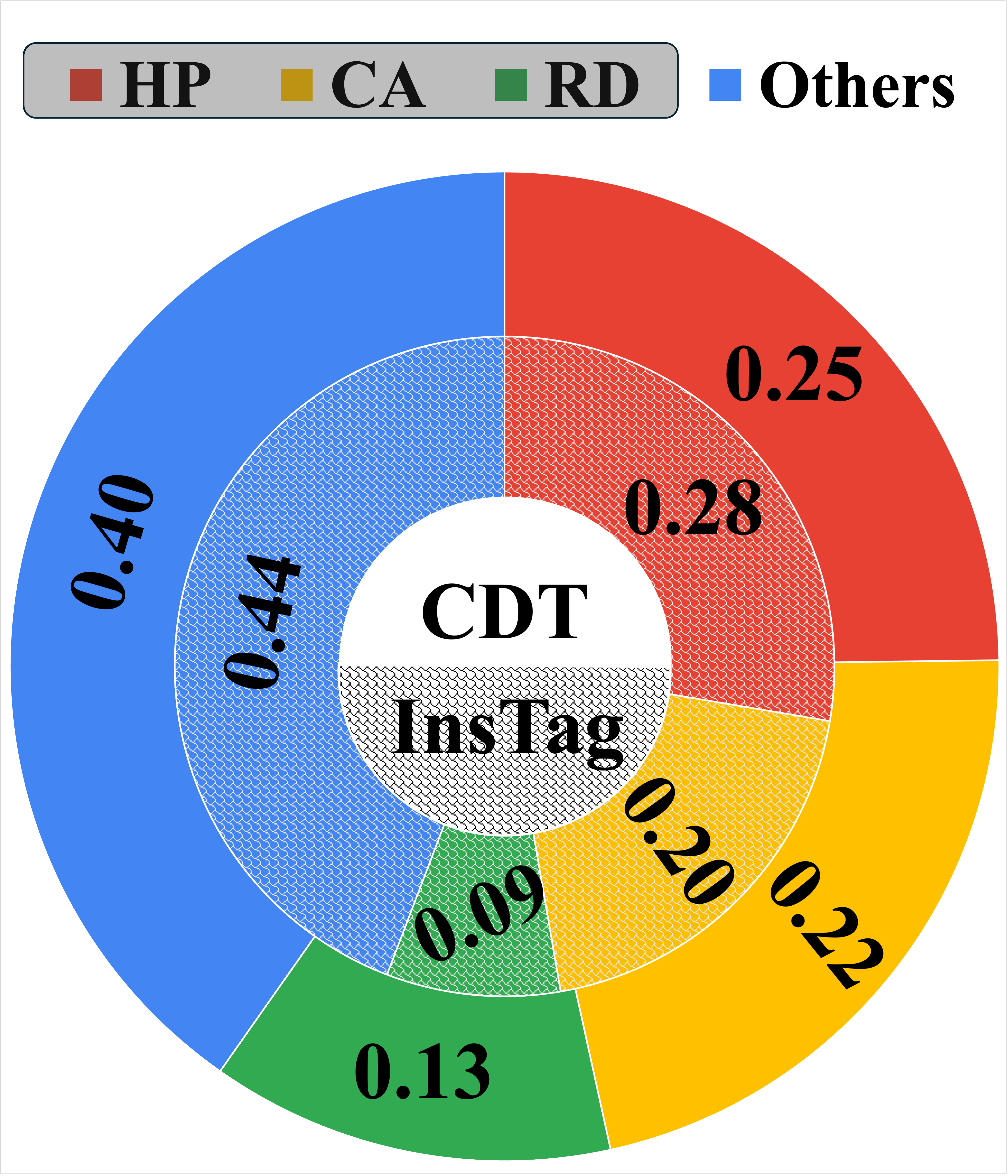}
        \label{fig:distribution_cognition}
	}
            \subfloat[Domain]{
        \includegraphics[width=0.3\linewidth]{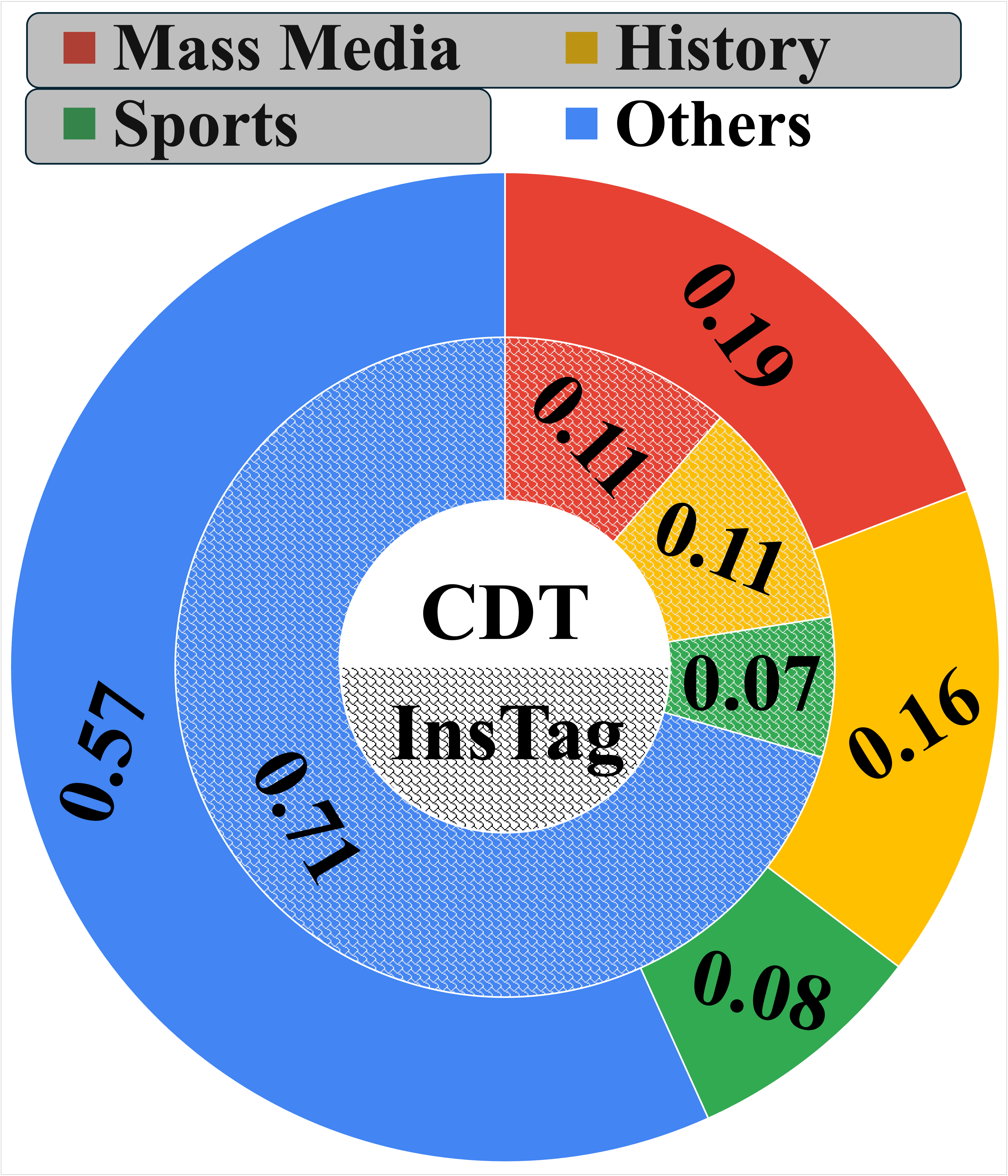}
        \label{fig:distribution_domain}
    }
    \subfloat[Task]{
        \includegraphics[width=0.3\linewidth]{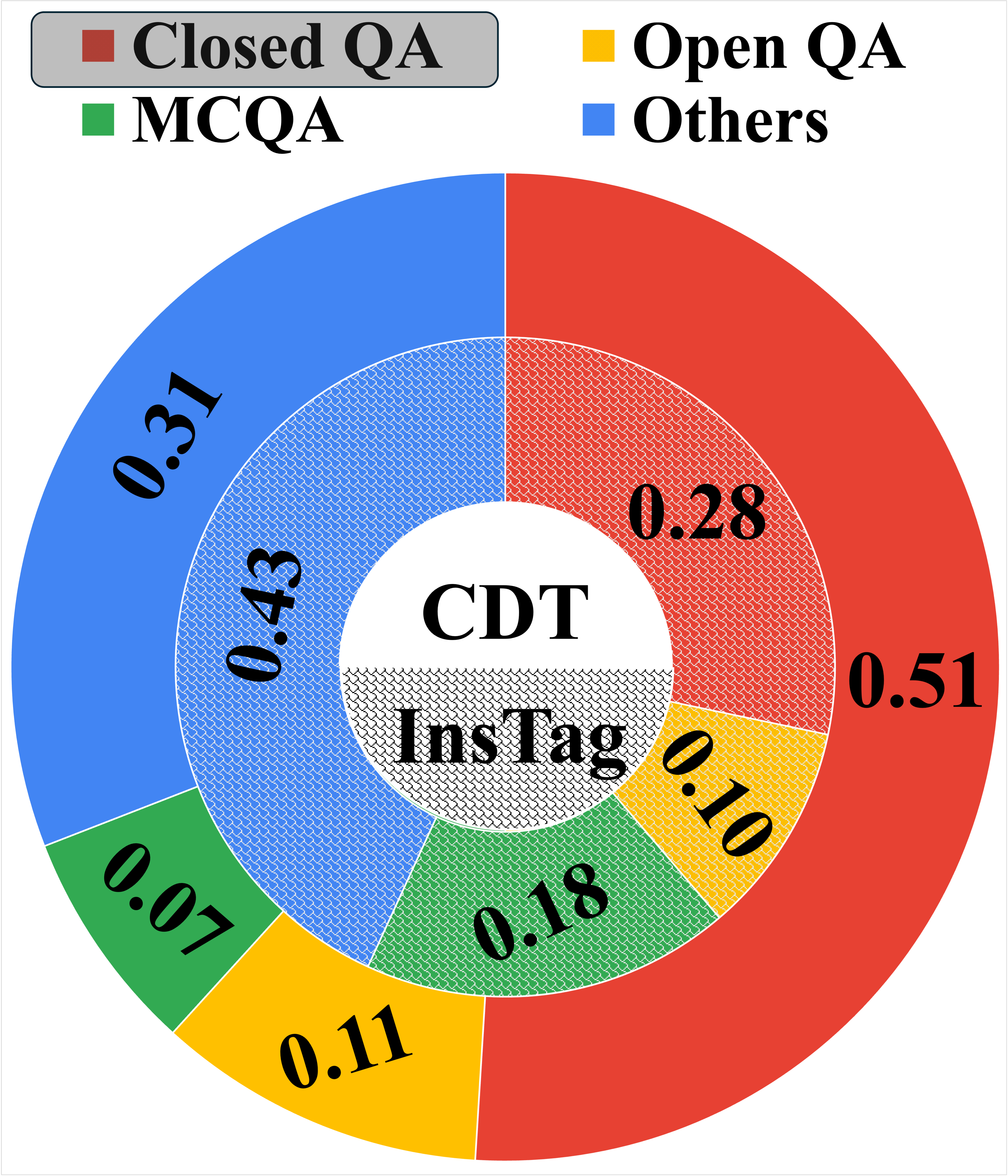}
        \label{fig:ditribution_task}
    }
    \caption{A comparison of the capability distributions of the data selected for the DROP test set using the CDT and INSTAG methods. The gray areas in the figure represent the capabilities required by DROP. MCQA stands for Multiple Choice QA.}
    \label{fig:distribution_all}
\end{figure}

\paragraph{Reasonability of Selected Data}
To further compare CDT with INSTAG, we use DROP as the targeted test set and analyze capability distributions by comparing the data selected by the INSTAG method with the tags annotated by CDT. 
As shown in Figure~\ref{fig:distribution_all}, CDT consistently aligns more closely with the target test set across the three dimensions. In cognition, both methods focus on key capabilities like HP (Hypothesis Generation), CA (Concept Abstraction), and RD (Reading Decoding). Although INSTAG shows a slightly higher distribution in the HP capability, CDT surpasses it in both CA and RD capabilities, demonstrating a more concentrated and dominant distribution. In the domain dimension, CDT selects substantially more data from relevant areas like history, sports, and mass media, which are central to the construction of DROP, as it prioritizes articles from these areas to support complex question generation. Regarding task dimension, CDT better captures the Closed QA capability, with a higher proportion of targeted samples. These results confirm that CDT effectively identifies and prioritizes the capabilities needed for the target test, thereby validating the strength and reliability of our framework.

\section{Conclusion}
In this work, we introduce the Cognition-Domain-Task (CDT) capability framework, offering a comprehensive and systematic approach to classify and decompose the capabilities of LLMs. By defining cognitive abilities based on Cattell-Horn-Carroll (CHC) theory and organizing domain and task capabilities into a structured taxonomy, we enable more nuanced categorization of LLM capabilities across various scenarios. Additionally, we trained a high-quality annotator on the Qwen2.5 model using the CDT framework.

We demonstrate the utility of CDT in two key applications. We first apply CDT to dataset evaluation using \texttt{Coverage} and \texttt{Balance} metrics to assess capability diversity and distribution. We then propose diversity-driven and capability-oriented data selection methods, both of which lead to substantial performance gains across multiple benchmarks. These results confirm the stability and effectiveness of CDT in guiding both dataset evaluation and data selection, highlighting the robustness and practical applicability of the framework.

\section*{Limitations}
Our method constructs a detailed three-dimensional LLM capability framework, CDT, and explores its application in dataset evaluation and data selection. However, there are still some limitations.

First, although the annotator trained on the Qwen-2.5 model achieves higher labeling accuracy across the three dimensions compared to INSTAG, there is still significant room for improvement. This could be addressed by adding more training data or incorporating specific knowledge from human experts to guide more accurate annotator training.

Second, when defining the three dimensions, we filter out multimodal capabilities, limiting the applicability of the CDT framework to a broader range of multimodal models. Future research could expand CDT to include relevant multimodal capability classifications and conduct experiments on multimodal models such as Qwen-VL~\citep{bai2023qwenvlversatilevisionlanguagemodel} and Llama-3.2~\cite{grattafiori2024llama3herdmodels}. 

Lastly, in our application of the CDT framework to LLMs, we have only explored its application in two scenarios. Future research may benefit from combining curriculum learning methods, such as Regmix~\citep{liu2024regmix}, with the CDT framework to dynamically adjust data distribution during training, potentially leading to even better results.

\section*{Acknowledgments}
This work was supported in part by Guangdong S\&T Program (Grant No. 2024B0101050003), Guangdong Basic and Applied Basic Research Foundation (Grant No. 2024A1515011491), and Shenzhen Science and Technology Program (Grant Nos. ZDSYS20230626091203008, KJZD20231023094700001, KQTD20240729102154066). 
Derek F. Wong was supported in part by the UM and UMDF (Grant Nos. MYRG-GRG2023-00006-FST-UMDF, MYRG-GRG2024-00165-FST-UMDF, EF2024-00185-FST).
We would like to thank the anonymous reviewers and meta-reviewer for their insightful suggestions.

\bibliography{anthology,custom}

\appendix
\section{Appendix}
\subsection{Design Rationale of CDT}
\label{app:CDTexplain}
This section elaborates on the foundational principles and design choices of the CDT framework. Its purpose is to provide deeper insight into the framework's application, particularly regarding the role of context and the relationship between the three capability dimensions.
\paragraph{Contextualized Instructions} A central design principle of the CDT framework is that it operates on entire instructions rather than isolated keywords or concepts. As a result, capability tagging is inherently context-dependent. This approach is critical for resolving the inherent ambiguity of language, since the capabilities required to interpret a term can vary substantially based on the instructional context. For example, consider the term \emph{“company”}, which can invoke different capabilities depending on the prompt. In the instruction \emph{``Suggest creative names for my new internet company,''} the context provided by \emph{``internet''} and \emph{``creative names''} indicates a Domain in Computer Science and a Cognition of Ideational Fluency. By contrast, in the instruction \emph{``Explain the legal definition of a limited liability company,''} the context of \emph{``legal definition''} shifts the relevant Domain to Law and the required Cognition to Concept Abstraction. This context-driven methodology enables the CDT framework to provide a precise and nuanced analysis of the capabilities demanded by each unique user instruction.

\paragraph{Relationship Between CDT Dimensions} The three dimensions of CDT are structured to be orthogonal, not hierarchical. They function as a multi-dimensional coordinate system. The Cognition axis describes how the model needs to reason or process information to fulfill the request, corresponding to how to think. The Domain axis identifies what the subject area or field of knowledge is, corresponding to what the topic is. The Task axis specifies what the user's explicit intent or the required output format is, corresponding to what to do. 

This structure ensures that a concept in one dimension is independent of the others. For example, a model might apply Pattern Recognition to perform a Detection task in the Literature domain by identifying the rhyme scheme in a poem, or use the same cognitive skill for a Generation task in the Economics  domain to summarize cyclical trends in market data.

\subsection{Cognitive Capability Construction}
\label{app:cogselect}
Starting with the ``narrow'' level of the CHC theory as defined in \citet{CHC}, we make systematic adaptations to tailor the taxonomy to the specific characteristics of LLMs, which differ from human cognition in both modality and operational dynamics. The adaptation process involves the following steps:
\begin{itemize}[leftmargin=1em]
\item \textbf{Exclusion of non-linguistic modalities} (e.g., speaking, listening, action, visual, olfactory abilities) since LLMs are text-based. This reduction brings the set from 82 to 33 abilities.
\item \textbf{Exclusion of non-core abilities} (e.g., memory and speed-related) that are less relevant to LLMs, as LLMs operate differently from humans in these aspects. We exclude abilities such as reading speed, writing speed, memory span, and others. This refinement reduces the set to 24.
\item \textbf{Exclusion of domain knowledge-related abilities}, as domain knowledge is a separate CDT dimension. We exclude abilities such as general information, lexical knowledge, geography achievement, and others. This step brings the number down to 13.
\item \textbf{Augmentation with LLM-relevant abilities}, such as logical analysis, abstract coding concepts, and problem decomposition, which are not emphasized in CHC but play a critical role in core LLM applications like code generation, reasoning, and instruction following. This increases the count to 16.
\item \textbf{Refinement of overlapping or broad definitions}, e.g., splitting Induction into pattern recognition, concept abstraction, and hypothesis generation. This final step results in a set of 18 distinct cognitive abilities.

\end{itemize}

\subsection{Capability Definition}
\label{app:tagdef}

\begin{table*}[!htb]
\centering
\scalebox{0.9}{
\begin{tabular}{p{4cm}p{2cm}p{10cm}}
\toprule
\textbf{Cognition}                                                                                    & \textbf{Abbreviation} & \textbf{Definition}                                                                                                                                         \\ \midrule
Pattern   Recognition                                                                                 & PR                    & Ability to identify recurring patterns, trends, or sequences within a given set of data or materials (e.g., detecting similarities in a sequence of numbers or text). \\ \hdashline
Concept   Abstraction                                                                                 & CA                    & Ability to form abstract concepts or categories based on shared characteristics or relationships among a set of materials.                              \\ \hdashline
Hypothesis   Generation                                                                               & HP                    & Ability to propose plausible explanations or predictions for incomplete information (e.g., inferring causes of a fictional conflict, suggesting scientific hypotheses). \\ \hdashline
General   Sequential Reasoning                                                                        & RG                    & Ability to start with stated   rules, premises, or conditions, and to engage in one or more steps to reach a   solution to a novel problem.                 \\ \hdashline
Quantitative   Reasoning                                                                              & RQ                    & Ability to inductively and   deductively reason with concepts involving mathematical relations and   properties.                                            \\ \hdashline
Reading   Decoding                                                                                    & RD                    & Ability to recognize and decode   words or pseudowords in reading.                                                                                          \\ \hdashline
Writing   Ability                                                                                     & WA                    & Ability to write with clarity of   thought, organization, and good sentence structure.                                                                      \\ \hdashline
Naming   Facility                                                                                     & NA                    & Ability to rapidly produce names   for concepts when presented with a text cue.                                                                             \\ \hdashline
Associational   Fluency                                                                               & FA                    & Ability to rapidly produce a   series of original or useful ideas related to a particular concept.                                                          \\ \hdashline
Expressional   Fluency                                                                                & FE                    & Ability to rapidly think of   different ways of expressing an idea.                                                                                         \\ \hdashline
Number     Facility                                                                                   & NM                    & Ability to rapidly and accurately manipulate and deal with numbers, from elementary skills of counting and recognizing numbers to advanced skills of adding, subtracting, multiplying, and dividing numbers.                                                                                                                       \\ \hdashline
Logical Analysis                                                                                   & LA                    & Ability to identify and apply logical structures, rules, and patterns within code or algorithms (e.g., recognizing logical constructs such as loops, conditions, or recursion in programming tasks).                                                                                                                       \\ \hdashline
Problem Decomposition                                                                                   & PD                    & Ability to systematically break down complex tasks into modular functional components, identify inter-component dependencies, and reconstruct solutions through controlled composition.                                                                                                                       \\ \hdashline
Abstract Coding Concept                                                                                   & AC                    & Ability to form abstract representations of programming concepts and apply them across different programming languages or environments (e.g., understanding concepts such as functions, variables, data structures, and algorithms in a generalized form, and applying them to solve problems in multiple programming languages).                                                                        \\ \hdashline
Sensitivity   to Problems/Alternative Solution  Fluency & SP                    & Ability to rapidly think of a   number of solutions to particular practical problem.                                                                        \\ \hdashline
Originality/   Creativity                                                                             & FO                    & Ability to rapidly produce   original, clever, and insightful responses (expressions, interpretations) to   a given topic, situation, or task.              \\ \hdashline
Ideational   Fluency                                                                                  & FI                    & Ability to rapidly produce a   series of ideas, words, or phrases related to a specific condition or object.   Quantity, not quality, is emphasized.        \\ \hdashline
Word   Fluency                                                                                        & FW                    & Ability to rapidly produce words   that have specific phonemic, structural, or orthographic characteristics   (independent of word meanings).               \\ \bottomrule
\end{tabular}}
\caption{The full definition of Cognition.}
\label{tab:cognition_definition}
\end{table*}

\begin{table*}[!htp]
\centering
\scalebox{0.9}{
\begin{tabular}{p{4cm}p{10cm}}
\toprule
\textbf{Domain} & \textbf{Sub-domain}                                               \\ \midrule
Language        & Linguistics,Literature,Multilingualism                            \\ \hdashline
Culture         & Tradition,Art,Sports,Mass Media,Music,Food                        \\ \hdashline 
Health          & Health                                                            \\ \hdashline
Natural Science & Biology,Earth Science,Astronomy,Chemistry,Physics                 \\ \hdashline
Math            & Mathematics,Logic                                                 \\ \hdashline
Social Science  & Economics,Law,Politics,Education,Sociology                        \\ \hdashline
Technology      & Agriculture,Computer   Science,Automation,Electronics,Engineering \\ \hdashline
Coding          & Coding                                                            \\ \hdashline
Humanities      & Communication,Religion,Philosophy,Ethics,History                  \\ \bottomrule
\end{tabular}}
\caption{The full definition of Domain.}
\label{tab:domain_definition}
\end{table*}

\begin{table*}[!htp]
\centering
\scalebox{0.9}{
\begin{tabular}{p{4cm}p{10cm}}
\toprule
Task                         & Definition                                                                                                                  \\ \midrule
Generation                   & Creating new information with human-input conditions,   involving the automatic generation of various text materials follow the instruction given by the user.\\ \hdashline
Rewrite                      & Taking a piece of text and rephrasing it while preserving its   original meaning, which may involve simplifying the language, changing the structure, or adjusting the tone. \\ \hdashline
Summarization                & Condensing longer texts into shorter versions while retaining   the key information and main ideas, making it easier to digest complex information. \\ \hdashline
Classification               & Assigning predefined labels or categories to text based on its   content, such as topic categorization.                     \\ \hdashline
Brainstorming                & Generating ideas, encouraging creative thinking, or exploring   possibilities.                                              \\ \hdashline
Sentiment                    & Determining the emotional tone or sentiment expressed in a   piece of text.                                                 \\ \hdashline
Completion                   & Continuing a given prompt with relevant and contextually   appropriate content, such as finishing sentences or filling in blanks. \\ \hdashline
Natural Language   Inference & Assessing the relationship between two sentences to determine   if one logically follows from the other (entailment), (contradiction), or if the relationship is unclear (neutral). \\ \hdashline
Bias and Fairness            & Evaluating models for potential bias, fairness, or harmfulness   in their outputs.                                          \\ \hdashline
Word Sense Disambiguation    & Determining which meaning of a word is used in a given   context, especially for words that have multiple meanings.         \\ \hdashline
Multiple Choice QA           & Answering questions by selecting the correct option from a   predefined set of possible answers based on provided information or context. \\ \hdashline
Closed QA               & Answering questions directly without access to external   knowledge.                                                        \\ \hdashline
Open QA                & Answering open-ended questions that can cover a wide range of topics, often without a single, definitive answer.                \\ \hdashline  
Extraction                & Identifying and extracting specific pieces of information from a given text.                \\ \hdashline  
Program Execution                & Executing or simulating the execution of a given program or script, processing inputs, performing operations, and returning outputs based on the specified instructions, often including code interpretation or debugging.                \\ \hdashline  
Detection                & Identifying the presence of specific elements, patterns, or anomalies in a given text, such as detecting spam or certain linguistic features like named entities or grammatical errors.               \\ \bottomrule
\end{tabular}}
\caption{The full definition of Task.}
\label{tab:task_definition}
\end{table*}

The detailed definitions and abbreviations for the cognition, domain, and task dimensions are provided in \Cref{tab:cognition_definition}, \Cref{tab:domain_definition}, and \Cref{tab:task_definition}, respectively. In defining the domain dimension, we first established the overarching domain and then carefully subdivided it into subdomains for labeling purposes.

\subsection{Capability Tagging Details}
\label{apx:tagging model}

\paragraph {Training Data} We collect 49K instruction samples from seven widely-used datasets: Selective Alpaca~\citep{liu2024selectit}, Dolly~\citep{DatabricksBlog2023DollyV2}, Open Assistant~\citep{NEURIPS2023_949f0f8f}, Super-Natural Instructions~\citep{wang2022supernaturalinstructionsgeneralizationdeclarativeinstructions}, Tulu 3~\citep{lambert2024t}, Flan V2~\citep{Longpre2023}, and WizardLM~\citep{xu2024wizardlm}. We randomly sample 7K queries per dataset and reserve 1K for testing.

\paragraph{Training Configuration} We fine-tune Qwen2.5-7B-Base for 1 epoch with a batch size of 32 and a cosine learning rate schedule initialized at 2e-5. 

\begin{figure*}[!htbp]
\centering
\subfloat[Cognition tagging prompt]{
\includegraphics[width=\linewidth]{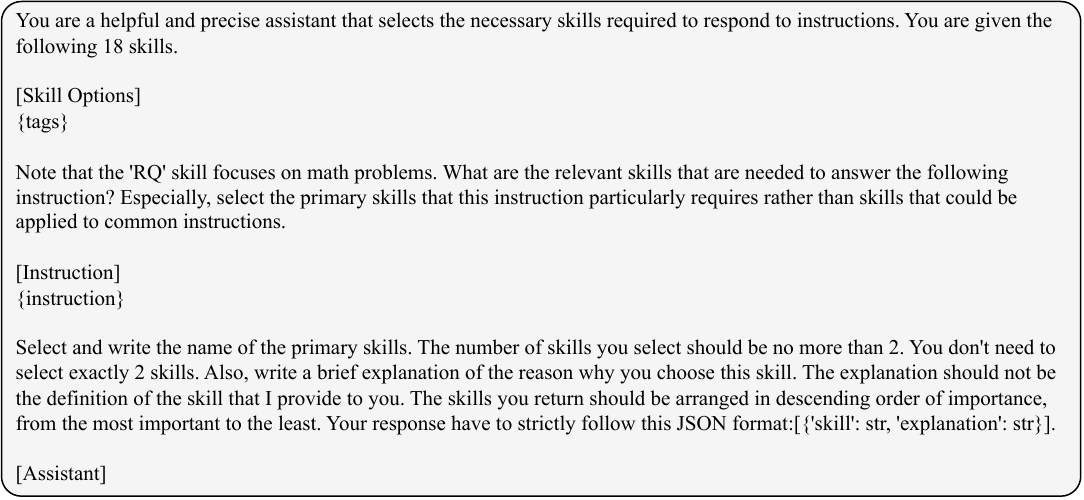}
\label{fig:prompt_cognition}
} \newline

\subfloat[Domain tagging prompt]{
\includegraphics[width=\linewidth]{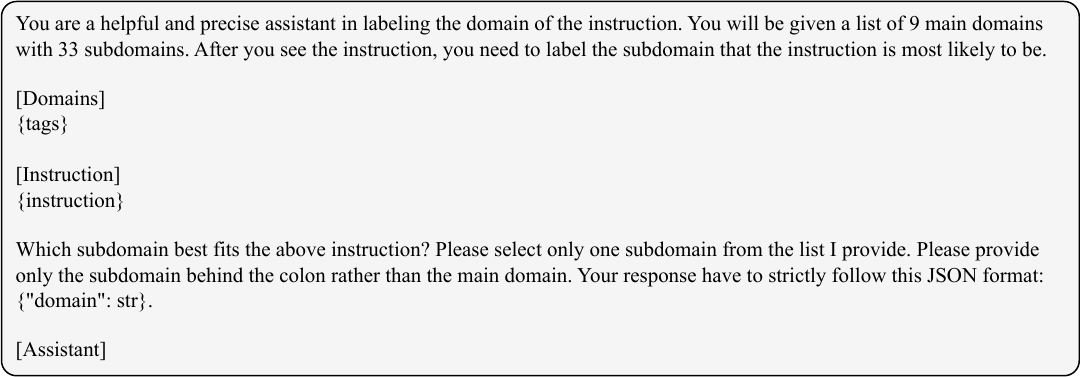}
\label{fig:prompt_domain}
} \newline

\subfloat[Task tagging prompt]{
\includegraphics[width=\linewidth]{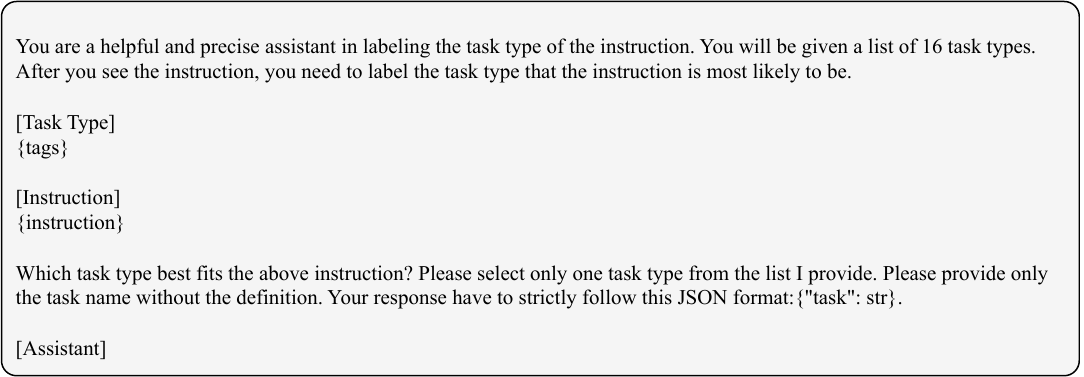}
\label{fig:prompt_task}
} \newline

\caption{The prompts we used on tagging.}
\label{apx:prompts}
\end{figure*}

\paragraph{Prompts} We design our prompts following the approaches proposed by \citet{lu2024instag, ye2024flaskfinegrainedlanguagemodel}. To mitigate position bias, we randomize the order of capabilities in the prompt for each data point. Additionally, when tagging cognitive capabilities, we ask the models to generate an explanation paired with each tag, as cognitive tasks require a deeper understanding of the instructions. All prompts are presented in \Cref{apx:prompts}. We concatenate the detailed descriptions of the query, tag, and instruction into a single input prompt. When labeling the cognition dimension, we restrict the model to output at most two tags, along with their corresponding explanations.

\paragraph {Human Evaluation} To evaluate the validity of the annotations generated by GPT-4o, we randomly selected 100 annotated entries from the training dataset to conduct a manual assessment. We evaluated the annotations based on the explanation of the labels in the cognition dimension, as well as the consistency of the task and domain dimensions with the original data, to determine whether GPT-4o’s annotations should be accepted as the ground truth.

We involved two human evaluators to assess the annotations, and the consistency of their assessments was as follows: cognition dimension consistency (95\%), domain dimension consistency (95\%),  task dimension consistency (85\%). Additionally, we calculated the average acceptance rates for GPT-4o’s annotations, which reflect the degree to which the evaluators agreed with GPT-4o’s judgments across each dimension: cognition (97.5\%), domain (87.5\%), task (87.5\%) and overall acceptance rate (90.5\%). From these results, we observe that GPT-4o’s annotations on the cognition dimension aligned more closely with human evaluations. This may be due to the fact that, in the cognition dimension, the cognitive abilities are more abstract and require a deeper understanding of the instructions. As a result, we not only require GPT-4o to output cognitive ability labels but also to provide explanations corresponding to these labels. 

The strong performance of GPT-4o validates the quality of our initial annotated dataset. However, relying on such a proprietary model for large-scale labeling of our 270K data pool is expensive and limits the broader adoption of the CDT framework. Therefore, to create a scalable and accessible solution, we use this high-quality dataset to train our own open-source capability annotators. The detailed cost-benefit analysis of this two-stage annotation approach is discussed below.

\paragraph {Annotation Cost Considerations} While the process of training the capability taggers requires an initial annotation phase, this process is largely automated using GPT-4o to generate fine-grained labels. This is a one-time, upfront investment designed not merely to label data for a single experiment, but to distill the nuanced definitions of our CDT framework into a set of efficient and open-source Qwen2.5-based annotators. The necessity of this approach is validated by our experiments: using the GPT-4o annotations as ground truth, we evaluate zero-shot performance on our test set. The Qwen2.5-7B-Base model achieves an average tagging accuracy of only 33.5\% across the three dimensions, while Qwen2.5-7B-Instruct reaches 53.5\%. In contrast, our annotators achieve an average accuracy of 85.1\%. This significant performance gap demonstrates that our initial data distillation is a crucial step for developing a reusable tagging tool. Once trained, our annotators can be applied to any number of future datasets at a low and predictable computational cost, far cheaper than repeated API calls to proprietary models, which makes the CDT framework highly scalable. As shown in \Cref{data selection exp}, this workflow strategically trades the upfront annotation investment for significant downstream efficiency: by precisely selecting data based on these capability labels, we achieve improved model performance using only a fraction of the total data, thereby substantially reducing the computational cost of the final fine-tuning phase.

\begin{table*}[!tp]
\centering
\vspace{-0.5em}
\renewcommand\tabcolsep{5.5pt}
\renewcommand\arraystretch{1.1}
\scalebox{0.9}{
\begin{tabular}{lcccccc}
\toprule
\textbf{Methods}        & \textbf{ARC-C} & \textbf{MMLU} & \textbf{BBH}  &  \textbf{CEVAL} & \textbf{TYDIQA} & \textbf{AVG.} \\ \midrule
\textbf{Base}    & 50.3           & \textbf{62.3}          & 57.6         & \textbf{46.8}           & 55.8            & 54.6          \\
\textbf{All}  & 52.1           & 60.6          & 56.2         & 46.3           & {\ul 57.3}            & 54.5          \\
\textbf{Random}  & 52.1           & 59.9          & {\ul 58.1}         & 46.4           & 57.0            & {\ul 54.7}          \\
\textbf{InsTag}  & \textbf{52.7}           & 60.5          & 56.0         & 46.3           & \textbf{57.7}            & 54.6          \\
\rowcolor[HTML]{C0C0C0} 
\textbf{CDT}     & {\ul 52.6}           & {\ul 61.5}          & \textbf{60.0}         & {\ul 46.6}           & 56.7            & \textbf{55.5} \\ \bottomrule
\end{tabular}}
\caption{Results of applying CDT in diversity-driven general data selection on the Mistral-7B-Base model.}
\label{table:mistral}
\end{table*}

\subsection{Data Selection Algorithm}
\label{apx:algorithm}

\begin{algorithm*}[!t]
\caption{Diversity-driven General Scenario Data Selection}
\label{appalg:general}
\KwData{$D^{'}_{pool}$: The capacity labeled data pool; $N$: Selection set size;}
\KwResult{$D_{train}$: The selected training dataset;}
\textbf{initialization}: $T_d$: All composite capabilities in the data pool; $D_{train} \gets \emptyset$\;
Sorting $T_d$ in descending order based on the number of corresponding data points in $D^{'}_{pool}$\;
\While{$|D_{train}| < N$}{
    $Flag \gets False$\;
    \For{each capability $f \in T_d$}{
        $D_f \gets Find\_Data(f, D^{'}_{pool})$\;
        \textcolor{ALG}{// Select data tagged with composite capability $f$ from $D^{'}_{pool}$}\\
        \If{$D_f \neq \emptyset$}{
             $d \gets Random(D_f,1)$\;
            \textcolor{ALG}{// Randomly select one data point from $D_f$}\\
            $D_{train} \gets \{d\} \cup D_{train}$\;
            $D^{'}_{pool} \gets D^{'}_{pool} \backslash \{d\} $\;
            $Flag \gets True$\;
        }
        \If{$|D_{train}| = N$}{
            \textbf{break}\;
        }
    }
    \If{$Flag = False$}{
         \textbf{break}\;
         \textcolor{ALG}{// All data points related to capability set $T_d$ are selected}\\
    }
}
\end{algorithm*}

\begin{algorithm*}[ht]
\caption{Capability-oriented Specific Scenario Data Selection}
\label{appalg:specific}
\KwData{$D^{'}_{pool}$: The capacity labeled data pool; $D^{'}_{valid}$: The capacity labeled validation set; $N$: Selection set size;}
\KwResult{$D_{train}$: The selected training dataset;}
\textbf{initialization}: $T_v$: Triplet capability set of validation set; $T^{*}_{v}$: Binary capability set; $T^{\star}_{v}$: Unary capability set; $D_{train} \gets \emptyset$\;

\For{each capability set $T \in \{T_v, T^{*}_{v}, T^{\star}_{v}\}$}{
    Sorting $T$ in descending order based on the number of corresponding data points in $D^{'}_{pool}$\;
    \While{$|D_{train}| < N$}{
        $Flag \gets False$\;
        \For{each capability $f \in T$}{
        \If{$N = |D_{train}|$}{
            \textbf{break}\;
        }
        $D_f \gets Find\_Data(f, D^{'}_{pool})$\;
        \textcolor{ALG}{// Select data tagged with composite capability $f$ from $D^{'}_{pool}$}\\
        \If{$D_f \neq \emptyset$}{
            $d \gets Random(D_f,1)$\;
            \textcolor{ALG}{// Randomly select one data point from $D_f$}\\
            $D_{train} \gets \{d\} \cup D_{train}$\;
            $D^{'}_{pool} \gets D^{'}_{pool} \backslash \{d\} $\;
            $Flag \gets True$\;
        }
        }
        \If{$Flag = False$}{
            \textbf{break}\;
            \textcolor{ALG}{// All data points related to capability set $T$ are selected}\\
        }
    }
}
\If{$|D_{train}| < N$}{
    \textcolor{ALG}{// Not enough data points labeled with the desired capabilities}\\
    $D_r \gets Random(D^{'}_{pool},N-|D_{train}|)$\;
    $D_{train} \gets D_r \cup D_{train}$\;
}
\end{algorithm*}

We present our diversity-driven general scenario data selection algorithm in Algorithm~\ref{appalg:general} and capability-oriented specific scenario in Algorithm~\ref{appalg:specific}.

\subsection{Experiments on Mistral Model}
\label{app:mistral}
In addition to the LLama2-7B-Base model, we also conducted experiments on the Mistral-7B-Base model, using 20\% of the training data in the general scenario as an example. The results are presented in~\Cref{table:mistral}. As shown, our method achieves the highest score of 55.5 among all baselines, further demonstrating the generalizability of our approach.

\end{document}